\documentclass{article}

\usepackage[final]{ProbNum25}

\usepackage{graphicx}
\usepackage{amssymb}
\usepackage{amsthm}
\usepackage{amsmath}
\usepackage{bm}
\usepackage{wrapfig}
\usepackage{physics}
\usepackage{booktabs}
\usepackage{colortbl}
\usepackage{xcolor}
\usepackage{makecell}
\usepackage{enumitem}
\usepackage{cancel}

\definecolor{darkblue}{RGB}{8,26,112}
\definecolor{goldenrod}{RGB}{218,165,32}
\definecolor{darkslateblue}{RGB}{72, 61, 139}
\definecolor{pinkorange}{rgb}{0.9882352941176471, 0.552941176470588,.3843137254901961}

\newtheorem{remark}{Remark}

\newcommand{\T}{\top}

\DeclareDocumentCommand\p{}{\opbraces{p}}
\DeclareDocumentCommand\N{}{\opbraces{\mathcal{N}}}
\DeclareDocumentCommand\E{}{\opbraces{\mathbb{E}}}
\DeclareDocumentCommand\V{}{\opbraces{\mathbb{V}}}

\DeclareDocumentCommand\g{}{\opbraces{g}}
\DeclareDocumentCommand\N{}{\opbraces{\mathcal{N}}}
\newcommand{\val}{y} 
\newcommand{\st}{x} 
\newcommand{\obs}{z} 
\newcommand{\proj}[1]{E_{#1}} %

\newcommand{\eye}[1]{\mathrm{I}_{#1}}

\newcommand\pcref[1]{(\cref{#1})}

\usepackage{amsmath,amsfonts,bm}

\def\eqref#1{equation~\ref{#1}}

\def\1{\bm{1}}

\DeclareMathAlphabet{\mathsfit}{\encodingdefault}{\sfdefault}{m}{sl}
\SetMathAlphabet{\mathsfit}{bold}{\encodingdefault}{\sfdefault}{bx}{n}

\newcommand{\R}{\mathbb{R}}

\usepackage[round]{natbib}

\usepackage[capitalise,nameinlink]{cleveref}

\probnumtitle{Propagating Model Uncertainty through Filtering-based Probabilistic Numerical ODE Solvers}
\probnumauthors{%
\name{Dingling Yao}%
\affiliation{Institute of Science and Technology Austria}%
\and%
\name{Filip Tronarp}%
\affiliation{Lund University}%
\and%
\name{Nathanael Bosch}%
\affiliation{Tübingen AI Center, University of Tübingen}%
}

\probnumabstract{
  Filtering-based probabilistic numerical solvers for ordinary differential equations (ODEs), also known as ODE filters, have been established as efficient methods for quantifying numerical uncertainty in the solution of ODEs.
  In practical applications, however, the underlying dynamical system often contains uncertain parameters, requiring the propagation of this model uncertainty to the ODE solution.
  In this paper, we demonstrate that ODE filters, despite their probabilistic nature, do not automatically solve this uncertainty propagation problem.
  To address this limitation, we present a novel approach that combines ODE filters with numerical quadrature to properly marginalize over uncertain parameters, while accounting for both parameter uncertainty and numerical solver uncertainty.
  Experiments across multiple dynamical systems demonstrate that the resulting uncertainty estimates closely match reference solutions.
  Notably, we show how the numerical uncertainty from the ODE solver can help prevent overconfidence in the propagated uncertainty estimates, especially when using larger step sizes.
  Our results illustrate that probabilistic numerical methods can effectively quantify both numerical and parametric uncertainty in dynamical systems.
}

\begin{document}

\section{Introduction}

Ordinary differential equations (ODEs) are used throughout the sciences to describe the evolution of dynamical systems over time.
They often appear in so-called initial value problems (IVP), given by equations
\begin{subequations}
  \label{eq:ivp}
  \begin{align}
    \dot{\val}(t) &= f_\theta(\val(t), t), \qquad t \in [0, T], \\
    \val(0) &= c_\theta,
  \end{align}
\end{subequations}
where
\(f_\theta : \mathbb{R}^d \times \mathbb{R} \to \mathbb{R}^d\) is a vector field that describes the dynamics of the system
and \(c_\theta \in \mathbb{R}^d\) is the initial condition,
both parameterized by some parameter \(\theta \in \mathbb{R}^e\).
The \emph{solution} of the IVP is a function \(\val_\theta : [0, T] \to \mathbb{R}^d\) that satisfies the ODE and the initial condition for the given parameter \(\theta\).

As the ODE solution \(\val_\theta\) is typically not available in closed form, it has to be approximated with one of many well-established numerical methods.
These include both classic, non-probabilistic approaches, in which the ODE solution is approximated by a single point estimate \( \hat{\val}_\theta(t) \)
(see e.g.\ \cite{hairer2008solving}),
as well as more recent probabilistic numerical methods, which return a posterior distribution over the ODE solution \(\p(\val(t) \mid \theta)\)
(see e.g.\  \cite{pnbook2022}).
The posterior in the latter approach captures the uncertainty in the ODE solution due to the numerical approximation of the solver, which for ODEs is typically a discretization of the time domain.
In this paper, we follow the probabilistic numerical perspective on ODE solvers, but we take an additional source of uncertainty into account,
namely uncertainty in the initial value problem itself.

In real-world applications, the initial value problem is rarely known exactly and both the initial value and the vector field parameters are often subject to uncertainty.
This uncertainty can be dealt with in two ways:
If measurements of the system trajectory are available, we can try to reduce the uncertainty and learn about the system by inferring the parameter \(\theta\) from data, for example via maximum likelihood estimation or Bayesian inference.
This is also known as the ``parameter inference'' problem, and it has been studied extensively over the years, both in non-probabilistic numerics
\citep{bard1974nonlinear,ramsay2007parameter},
and in the context of probabilistic numerical ODE solvers
\citep{kersting20_differ_likel_fast_inver_free_dynam_system,tronarp2022fenrir,dalton2024}.
On the other hand, if no data is available, we can compute a distribution over possible system trajectories of the uncertain system by propagating the uncertainty from the system parameters to the ODE solution itself.
This is the problem of interest in this paper, and we refer to it as the ``uncertainty propagation'' problem.

Previous work on uncertainty propagation mainly resides in the non-probabilistic numerics literature.
\citet{gerlach2020} consider the uncertainty propagation problem from a Koopman operator perspective, and propagate uncertainty from the initial value to the solution of the ODE by combining ODE solvers with numerical quadrature.
Uncertainty propagation also plays a role in certain machine learning problems where the ODE vector field is learned from data, e.g. with a Gaussian process \citep{ridderbusch2023past}.
The problem also relates to the broader field of \emph{random ODEs}, where the ODE vector field is itself a random process, but this is a much more general setting than the one we consider here
\citep{RODEbook2017}.
In the probabilistic numerics literature, \citet{pfoertner2023physicsinformed} have propagated uncertainty through linear PDEs with Gaussian-process-based methods.
On nonlinear ODEs on the other hand, filtering-based approaches have been established as an efficient class of methods
\citep{kersting18convrates,schober16_probab_model_numer_solut_initial_value_probl,tronarp18_probab_solut_to_ordin_differ}---but to the best of our knowledge, uncertainty propagation with these methods has not been studied in detail yet.


In this paper, we present a novel approach for uncertainty propagation with probabilistic numerical ODE solvers.
We formalize the uncertainty propagation problem in the context of probabilistic numerical ODE solvers and show how existing methods are not able to propagate uncertainty out-of-the-box---despite them being \emph{probabilistic}.
We then propose a new approach to uncertainty propagation by combining filtering-based ODE solvers with numerical quadrature.
We experimentally validate our method on multiple dynamical systems and show how the resulting uncertainty estimates closely match the true propagated uncertainty.
We also
show how the numerical uncertainty of the ODE solver enters the total propagated uncertainty and how it can prevent overconfidence compared to non-probabilistic ODE solvers.

\section{Filtering-based Probabilistic Numerical ODE Solvers}
\label{sec:pn-ode}
In this section we provide a brief introduction to filtering-based probabilistic numerical ODE solvers, also known as \emph{ODE filters}.
Given an ODE-IVP as in \cref{eq:ivp}
together with a known, fixed parameter vector $\theta$,
we aim to compute posterior distributions of the form
\begin{equation}
  \label{eq:posterior:bg}
  \p( \val(t) \mid \val(0) = c_\theta, \{\dot{\val}(t_n) = f_\theta(\val(t_n), t_n)\}_{n=1}^N ),
\end{equation}
where $\{t_n\}_{n=1}^N \subset [0, T]$ is a chosen discretization of the time domain.
We refer to this quantity, and approximations thereof, as the \emph{probabilistic numerical solution} of the IVP.

In a nutshell, ODE filters approach this problem by posing it as a Gauss--Markov regression problem and solving it with Bayesian filtering and smoothing methods.
In the following, we briefly introduce the building blocks of ODE filters, namely the Gauss--Markov process prior, the observation models, and the resulting discrete-time inference problem;
for a more complete introduction to the topic, refer to \citet{tronarp18_probab_solut_to_ordin_differ}.
We then briefly discuss the current limitations of these methods for uncertainty propagation.

\subsection{Gauss--Markov process prior}
ODE filters model the ODE solution \(\val(t)\) as a Gauss--Markov process, defined as the output of a linear time-invariant stochastic differential equation (LTI-SDE)
\begin{subequations}
  \label{eq:prior:sde}
  \begin{align}
    \st(0) &\sim \N(\st(0); \mu^-_0, \Sigma^-_0), \\
    \dd{\st(t)} &= F \st(t) \dd{t} + \Gamma^{1/2} \dd{w(t)}, \\
    \val(t) &= \proj{0} \st(t), \\
    \dot{\val}(t) &= \proj{1} \st(t),
  \end{align}
\end{subequations}
where \(\st(t) \in \R^D\) is the \emph{state} of the process at time \(t\) of dimension \(D\),
\(\mu_0 \in \R^D\) is the initial state mean,
\(\Sigma_0 \in \R^{D \times D}\) is the initial state covariance,
\(F \in \R^{D \times D}\) is the drift matrix,
\(\Gamma^{1/2} \in \R^{D \times d}\) is the dispersion matrix,
\(w: \R_{\geq 0} \to \R^d\) is a standard Wiener process,
and \(\proj{0}, \proj{1} \in \R^{d \times D}\) are projection matrices that map the state to the ODE solution and its derivative, respectively.

In discrete time, LTI-SDEs satisfy exactly discrete-time transition densities of the form
\begin{equation}
  \st(t+h) \mid \st(t) \sim \N( \st(t+h); \Phi(h) \st(t), Q(h) ),
\end{equation}
where \(\Phi(h), Q(h)\) can be derived from the drift and dispersion matrices \(F, \Gamma\)
\citep{sarkka_solin_2019}.

In this paper, we consider ODE filters that model the solution with a \(q\)-times integrated Wiener process prior, which has transition and covariance matrices given by
\citep{kersting18convrates}
\begin{subequations}
\begin{align}
  \Phi(h) &= \eye{d} \otimes \breve{\Phi}(h), \\
  Q(h) &= \eye{d} \otimes \kappa^2 \breve{Q}(h),
\end{align}
\end{subequations}
where
\begin{subequations}
  \begin{align}
    \left[ \breve{\Phi}(h) \right]_{ij} &= \mathbb{I}_{i \leq j} \frac{h^{j-1}}{(j-i)!}, \\
    \left[ \breve{Q}(h) \right]_{ij} &= \frac{h^{2q+1-i-j}}{(2q+1-i-j)(q-i)!(q-j)!},
  \end{align}
\end{subequations}
leading to a state dimension of \(D = d \cdot (q + 1)\).
The \emph{diffusion parameter} \(\kappa^2 \in \R_{>0}\)
will typically be estimated by the ODE filter via quasi-maximum likelihood estimation
(see e.g. \cite{tronarp18_probab_solut_to_ordin_differ,bosch20_calib_adapt_probab_ode_solver}).
The solution \(\val(t)\) and its derivative \(\dot{\val}(t)\) can be extracted from the state \(\st(t)\) via multiplication with the projection matrices
\begin{subequations}
\begin{align}
  \proj{0} = \eye{d} \otimes \mqty[1 & 0 & 0 & \cdots & 0]^\T, \\
  \proj{1} = \eye{d} \otimes \mqty[0 & 1 & 0 & \cdots & 0]^\T.
\end{align}
\end{subequations}
But other Gauss--Markov priors are also suitable for ODE filtering, such as the
integrated Ornstein--Uhlenbeck processes or Matérn processes.
Refer to \citet{tronarp20_bayes_ode_solver} or \citet{bosch2023probabilistic} for more details.

\subsection{IVP information operators}
To relate the Gauss--Markov prior to the solution of the IVP, we need to enforce both the initial value and the ODE itself.
In ODE filters, this is done by introducing two observation models:
\begin{align}
  \obs_0 \mid \st(0) &\sim \delta \left( \proj{0} \st(0) - c_\theta \right), \\
  \obs(t) \mid \st(t) &\sim \delta \left( \proj{1} \st(t) - f \left( \proj{0} \st(t), t, \theta \right) \right).
\end{align}
If \(\st(t)\) satisfies the IVP, then we have
\(\obs_0 = 0\) as well as \(\obs(t) = 0\) for all \(t \in [0, T]\).
Thus, conversely, if we were to condition \(\st(t)\) on \(\obs_0 = 0\) and \(\obs(t) = 0\), then the resulting process would satisfy the IVP.
Unfortunately this posterior is not tractable, but we will show how to approximate it in the next section.

\begin{remark}[Initialization with higher-order derivatives]
  For improved accuracy and numerical stability, the state can also be initialized on higher-order derivatives at the initial value.
  These can be computed efficiently via Taylor-mode automatic differentiation
  \citep{kraemer20_stabl_implem_probab_ode_solver},
  and we can then extend the initial observation model to include these higher-order derivatives.
  In our implementation, we use this strategy and initialize the solver on all derivatives up to order \(q\).
\end{remark}

\subsection{Probabilistic numerical ODE solutions as Bayesian filtering and smoothing}

\begin{figure*}[t]
  \begin{center}
    \includegraphics[width=0.9\linewidth]{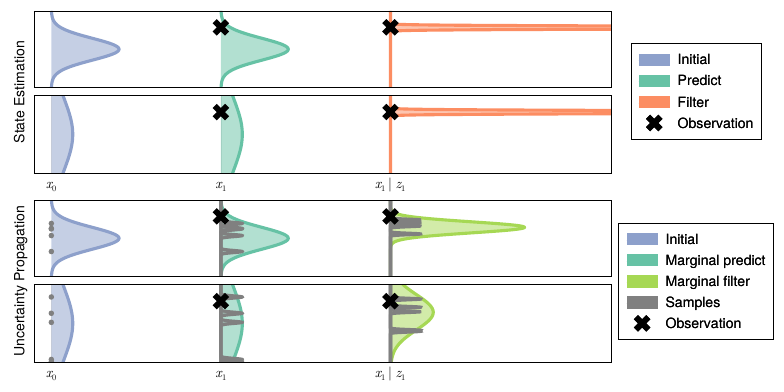}
  \end{center}
  \caption{
    \emph{Difference between state estimation and uncertainty propagation.}
    We consider two simple linear state-space models,
    both with
    transition model
    \(\p(x_1 \mid x_0) = \N(x_0, 0.01)\),
    observation model
    \(\p(y_1 \mid x_1) = \N(x_0, 0.01)\),
    and observation \(y_1 = 2\),
    and with two different initial distributions
    \(\p(x_0) = \N(0, 1)\) and \(\p(x_0) = \N(0, 10)\).
    The top rows show the distributions computed by a regular Kalman filter applied to these models, which solves the corrensponding \emph{state estimation} problems.
    The bottom rows instead show the solutions to the \emph{uncertainty propagation} problem, in which the initial state should not be inferred but marginalized out.
    This aspect is visualized by the samples in gray, which are drawn from the initial distribution, propagated forward by running a separate Kalman filter for each sample, and then aggregated.
    In addition, the figure also demonstrates how the filter distribution barely changes when the initial uncertainty is increased, whereas the marginal filter distribution becomes significantly wider.
  }
  \label{fig:stateestimation_vs_uncertaintypropagation}
\end{figure*}

Instead of attempting to satisfy the ODE everywhere, we discretize the time domain into a grid \(\{t_n\}_{n=1}^N \subset [0, T]\) and only condition the process on ODE observations at these time points.
This leads to the following inference problem:
\begin{subequations}
  \label{eq:inference-problem}
  \begin{align}
    \st(0) &\sim \N( \st(0); \mu_0^-, \Sigma_0^- ), \\
    \st(t_{n+1}) \mid \st(t_n) &\sim \N( \Phi(h_n) \st(t_n), Q(h_n) ), \\
    \obs_0 \mid \st(0) &\sim \delta \left( \proj{0} \st(0) - c_\theta \right), \\
    \obs(t_n) \mid \st(t_n) &\sim \delta \left( \proj{1} \st(t_n) - f( \proj{0} \st(t_n), t_n, \theta) \right),
  \end{align}
\end{subequations}
where \(h_n = t_{n+1} - t_n\) is the step size for the \(n\)-th time step,
and with observations
\(\obs_0 = 0\) and \(\obs(t_n) = 0\) for all \(n\).
This problem is known as a nonlinear Gauss--Markov regression problem, and its solution can be approximated efficiently with Bayesian filtering and smoothing,
for instance with an extended Rauch--Tung--Striebel smoother.
This yields a Gaussian process posterior with marginals
\begin{equation}
  \p( \st(t) \mid \obs_0=0, \{\obs(t_n) = 0\}_{n=1}^N, \theta ) \approx \N( \mu_\theta(t), \Sigma_\theta(t) ).
\end{equation}
or more compactly,
\begin{equation}
  \label{eq:posterior}
  \p( \st(t) \mid \mathcal{D}_\text{PN}, \theta ) \approx \N( \mu_\theta(t), \Sigma_\theta(t) ),
\end{equation}
where \(\mathcal{D}_\text{PN}\) denotes the data.
The posterior ODE solution
\(\p( \val(t) \mid \mathcal{D}_\text{PN}, \theta )\)
can then be extracted from \cref{eq:posterior} via projection with \(\proj{0}\);
this provides an approximation to \cref{eq:posterior:bg} and is thus a \emph{probabilistic numerical solution} of the IVP, computed by the ODE filter.

\subsection{Can ODE filters propagate Gaussian uncertainty out of the box?}
The ODE filters we introduced above formulate ``ODE solving'' as probabilistic inference.
But, contrary to what one might think at first, these methods do not automatically also solve the uncertainty propagation problem of interest:
The posterior computed by the ODE filter, shown in \cref{eq:posterior}, conditionally depends on the parameter \(\theta\), but the desired posterior in the uncertainty propagation problem of interest requires its marginalization, i.e.
\( \p( \val(t) \mid \mathcal{D}_\text{PN} ) = \int \p( \val(t) \mid \mathcal{D}_\text{PN}, \theta ) \p(\theta) \dd{\theta} \).
Even in the case in which only the initial value of the IVP is uncertain this additional marginalization is necessary;
\cref{fig:stateestimation_vs_uncertaintypropagation} demonstrates this visually with a simple linear example.
In the following, we will show why this is generally the case and how Baysian filters do not \emph{propagate uncertainty} but rather \emph{estimate states}.

To this end, we consider a generic state-space model with only a single time step:
\begin{subequations}
\begin{align}
  \text{Initial state:} \qquad & \p(\st_0), \\
  \text{Transition model:} \qquad & \p( \st_1 \mid \st_0 ), \\
  \text{Observation model:} \qquad & \p( \obs_1 \mid \st_1 ).
\end{align}
\end{subequations}
Our goal here is now to compute the distribution over the state \(\st_1\) given the observation \(\obs_1\), while also propagating the uncertainty from the initial state \(\st_0\) to the state \(\st_1\).
That is, we want to compute
\begin{equation}
  p_\text{goal}( \st_1 \mid \obs_1 )
  = \int \p( \st_1 \mid \obs_1, \st_0 ) \p( \st_0 ) \dd{\st_0}.
\end{equation}
To relate this to the given state-space model, we can expand the conditional probability with Bayes' rule and obtain
\begin{equation}
  \label{eq:up-example:goal}
  p_\text{goal}( \st_1 \mid \obs_1 )
  = \int
    \frac{
    \p( \obs_1 \mid \st_1 )\
    \p( \st_1 \mid \st_0 )
    }{
    \p( \obs_1 \mid \st_0 )
    }
    \ \p( \st_0 ) \dd{\st_0}.
\end{equation}
Now if we were to simply apply a generic Bayesian filter to this problem
(e.g. described by \cite{sarkka_bayesianfilteringandsmoothing}),
we would first compute the \emph{predict} step and obtain
\begin{equation}
  p_\text{predict}( \st_1 ) = \int \p( \st_1 \mid \st_0 ) \p( \st_0 ) \dd{\st_0},
\end{equation}
and then copmute the \emph{update} step which yields
\begin{equation}
  p_\text{filter}( \st_1 \mid \obs_1 )
  = \frac{ \p( \obs_1 \mid \st_1 )\  p_\text{predict}( \st_1 ) }{ \p( \obs_1 ) };
\end{equation}
or altogether, we compute
\begin{equation}
  \label{eq:up-example:filter}
  p_\text{filter}( \st_1 \mid \obs_1 )
  = \int \frac{ \p( \obs_1 \mid \st_1 )\  \p( \st_1 \mid \st_0 ) }{ \p( \obs_1 ) }
    \ \p( \st_0 ) \dd{\st_0}.
\end{equation}
Comparing \cref{eq:up-example:filter} to \cref{eq:up-example:goal}, we see that
the normalization constant differs between the two expressions and thus both quantities differ, i.e. \(p_\text{filter} \neq p_\text{goal}\).
Thus, a general Bayesian filter does not actually propagate the uncertainty from the initial state forward in time, but rather reduces the uncertainty by estimating the state from the observations at each time step---%
or in short, it performs \emph{state estimation} and not \emph{uncertainty propagation}.
\Cref{fig:stateestimation_vs_uncertaintypropagation} also visualizes this difference with a simple linear example.
Overall, this shows the need for a modification or extension of the existing ODE filtering algorithm to perform uncertainty propagation.

\section{Propagating Uncertainty through ODE Filters}
\label{sec:theory}

This section introduces the uncertainty propagation problem of interest and presents a general approach to approximate its solution with probabilistic numerical ODE solvers.
Consider an initial value problem (IVP) as given in \cref{eq:ivp} and let \(\p(\theta)\) be a distribution over the system parameter \(\theta\).
Our goal in uncertainty propagation is then to compute the expected value and variance of the ODE solution \(\val_\theta(t)\) with respect to the parameter distribution \(\p(\theta)\), which correspond to expectations of the form
\begin{equation}
  \label{eq:up:goal}
  \E[ \g( \val_\theta(t) ) ]_{\p(\theta)} = \int \g( \val_\theta(t) ) \p(\theta) \dd{\theta}
\end{equation}
with
\( \g(\val) = \val \)
and
\( \g(\val) = (\val - \E[\val])^2 \)
for the expectation and variance, respectively,
for any chosen time point \(t \in [0, T]\) (or a collection thereof).

As the true ODE solution \(\val_\theta(t)\) is not available in closed form,
we approximate it with the probabilistic numerical ODE solution
\(\p(\val(t) \mid \theta)\)
and compute its moments instead, while also marginalizing out the numerical uncertainty of the ODE solver itself.
This yields
\begin{equation}
  \label{eq:posterior:int}
  \E[ \g( \val_\theta(t) ) ]_{\p(\theta)} \approx \int \int \g( \val(t) ) \p(\val(t) \mid \theta)
  \dd{\val(t)}
  \p(\theta)
  \dd{\theta}.
\end{equation}
\Cref{eq:posterior:int} now involves only known quantities, but the nested integrals are in general intractable.
In this section, we present our two-step approach to approximate these integrals via numerical integration.

Re-ordering the terms in \cref{eq:posterior:int}, we obtain
\begin{equation}
  \E[ \g( \val_\theta(t) ) ]_{\p(\theta)} \approx \\
  \int \g( \val(t) )
  \left(
    \int \p(\val(t) \mid \theta) \p(\theta) \dd{\theta}
  \right)
  \dd{\val(t)}.
\end{equation}
Our proposed two-step approach now proceeds as follows:
First, we approximate the marginal PN solution
\begin{equation}
\p(\val(t)) = \int \p( \val(t) \mid \theta ) \p(\theta) \dd{\theta}
\end{equation}
with numerical integration; this yields a Gaussian mixture distribution.
Second, we compute the expectation and variance of the Gaussian mixture distribution and obtain the desired moments of the ODE solution posterior.
%
We will discuss each of these steps in more detail in the following.


\subsection{Approximating the ODE solution posterior with numerical integration}
\label{sec:integration}
As the parameter \(\theta\) enters the ODE solution \(\p( \val(t) \mid \theta)\) in a nonlinear way, the integral
\begin{equation}
  \p(\val(t)) = \int \p( \val(t) \mid \theta ) \p(\theta) \dd{\theta}
\end{equation}
is again intractable.
One common way to approximate intractable integrals is with numerical quadrature.
In a nutshell, a numerical quadrature scheme approximates an integral by a weighted sum of evaluations of the integrand at a finite set of nodes, i.e.
\begin{equation}
  \int \p( \val(t) \mid \theta ) \p(\theta) \dd{\theta} \approx \sum_{i=1}^N w_i \cdot \p( \val(t) \mid \theta_i ),
\end{equation}
with \emph{nodes} \(\theta_i \in \R^e\) and weights \(w_i \in \R\).
These nodes and weights are provided by the chosen numerical integration scheme, and the quality of the approximation depends on the number of nodes and the choice of the nodes and weights.

In this paper, we focus on Gaussian parameter distributions \(\p(\theta) = \N(\mu_\theta, \Sigma_\theta)\) and
mainly consider the \emph{third-degree spherical cubature} scheme
\citep{genz1999stochastic,lu2004higher}, given by
\begin{subequations}
\label{eq:spherical-cubature}
\begin{align}
  \theta_i &= \begin{cases}
    \mu_\theta + \sqrt{d} \Sigma_\theta^{1/2} e_i, & \text{for } i = 1, \ldots, d, \\
    \mu_\theta - \sqrt{d} \Sigma_\theta^{1/2} e_{i-d}, & \text{for } i = d+1, \ldots, 2d,
  \end{cases} \\
  w_i &= \frac{1}{2d}, \!\ \qquad\quad\qquad\quad\quad\quad \text{for } i = 1, \ldots, 2d,
\end{align}
\end{subequations}
where
\(d\) is the dimension of the parameter space,
\(e_i\) are the canonical basis vectors in \(\R^d\),
and where \(\Sigma_\theta^{1/2}\) denotes the left Cholesky factor of \(\Sigma_\theta\), i.e. \(\Sigma_\theta = \Sigma_\theta^{1/2} (\Sigma_\theta^{1/2})^{\T}\).
This scheme is well-suited for Gaussian parameter distributions \(\p(\theta)\) and is known to be quite efficient in moderate to high nonlinear systems \citep{sarkka_bayesianfilteringandsmoothing}.
But other numerical quadrature methods could also be used, as discussed in \cref{remark:numerical-integration}.

\begin{remark}[Other numerical quadrature methods]
  \label{remark:numerical-integration}
  \ \\
 Choosing a suitable quadrature method typically depends on the integrand and the parameter distribution.
  For Gaussian parameter distributions, \emph{sigma point methods} such as spherical cubature, the unscented transform \citep{julier1997new}, or Gauss--Hermite quadrature \citep{abramowitz1968handbook} are particularly well-suited.
  For non-Gaussian parameter distributions,
  \emph{generalized Gaussian quadrature rules} can be used to approximate the integral, such as  Gauss--Legendre quadrature for uniform distributions or Gauss--Laguerre quadrature for exponential distributions
  \citep{golub1969calculation,stroud1966gaussian}.
  For more general distributions, if the functional form of $p(\theta)$ is known, one can approximate the integral by combining Gauss--Hermite Quadrature with importance sampling (adaptive GHQ).
  Otherwise, Monte Carlo integration simply averages the integrand over samples from the distribution and is thus a general-purpose method that can be used for any distribution, but is typically less efficient than quadrature methods.
\end{remark}

By applying the chosen numerical quadrature scheme to the ODE solution posterior \(\p(\val(t))\), we obtain a weighted sum of the form
\begin{equation}
  \p(\val(t)) \approx \sum_{i=1}^N w_i \p( \val(t) \mid \theta_i ),
\end{equation}
with \(w_i, \theta_i\) as in \cref{eq:spherical-cubature}.
Since each individual probabilistic numerical ODE solution is a Gaussian process posterior with Gaussian marginals
\(\p( \val(t) \mid \theta_i ) = \N( \mu_i(t), \Sigma_i(t) )\),
the mixture distribution \(\p(\val(t))\) is a Gaussian mixture distribution of the form
\begin{equation}
  \label{eq:mixture}
  \p(\val(t)) \approx \sum_{i=1}^N w_i \ \N( \mu_i(t), \Sigma_i(t) ).
\end{equation}

\subsection{Computing the expectation and covariance of the ODE solution posterior}
Now that we approximated the ODE solution posterior with a Gaussian mixture distribution as in \cref{eq:mixture},
we can compute its expectation and variance in closed form:
The mean of a mixture distribution
is given by the weighted sum of the means of the mixture components, i.e.
\begin{equation}
  \label{eq:mixture_expectation}
  \E[ \val(t) ]_{\p(\val(t))} = \sum_{i=1}^N w_i \mu_i(t),
\end{equation}
and its covariance is given by
\begin{align}
  \label{eq:mixture_variance}
  \begin{split}
  \V[ \val(t) ]_{\p(\val(t))}
  &=
  \\ \sum_{i=1}^N w_i &\left[ \Sigma_i(t) + \left( \mu_i(t) - \bar{\mu}(t) \right) \left( \mu_i(t) - \bar{\mu}(t) \right)^\T \right],
  \end{split}
\end{align}
where \(\mu_i(t), \Sigma_i(t)\) are the mean and covariance of the \(i\)-th mixture Gaussian component at time \(t\), respectively,
and with \(\bar{\mu}(t) = \E[ \val(t) ]_{\p(\val(t))}\) as in \cref{eq:mixture_expectation}.
This yields the desired moments of the ODE solution with respect to the parameter distribution \(\p(\theta)\) and the numerical uncertainty of the ODE solver itself and concludes the algorithm.

\begin{remark}[The role of numerical uncertainty and relation to non-probabilistic numerical simulators]
  \label{remark:pn-non-pn}
  The proposed algorithm can also be applied to non-probabilistic ODE solvers.
  If we interpret non-probabilistic numerical solvers as solvers that return a Dirac distribution
  \(\p(\val(t) \mid \theta) = \delta\!\left( \val(t) - \hat{\val}_\theta(t) \right)\),
  the derivations above still hold,
  and the only difference is that the term in the covariance, that corresponds to the numerical uncertainty of the ODE solver, vanishes:
  \begin{align}
    \begin{split}
      \V[ \val(t) ]_{\p(\val(t))}
      &=
      \\
      \sum_{i=1}^N w_i &\left[ \cancel{\Sigma_i(t)} + \left( \mu_i(t) - \bar{\mu}(t) \right) \left( \mu_i(t) - \bar{\mu}(t) \right)^\T \right],
    \end{split}
  \end{align}
  These results correspond exactly to the uncertainty propagation through non-probabilistic numerical solvers approach proposed by \citet{gerlach2020} (for our particular choice of quadrature scheme).
  We will expore this aspect and the utility of the ODE solver's uncertainty further in \cref{sec:experiments:why_pn}.
\end{remark}

\section{Experiments}
\label{sec:experiments}
This section evaluates the proposed probabilistic uncertainty propagation method on a range of dynamical systems.
We first validate the proposed uncertainty propagation method on multiple dynamical systems and qualitatively compare the resulting mean and variance estimates to an accurate reference solution
\pcref{sec:experiments:examples}.
We then investigate the utility of the numerical uncertainty estimates provided by the underlying ODE solver by disentangling the resulting numerical uncertainty into PN and non-PN parts
\pcref{sec:experiments:why_pn}.

\paragraph{Implementation details.}
In our implementation we build on the ODE filters provided by the ProbNum python package
\citep{wenger2021probnum},
which follow a number of practices for numerically stable implementation
\citep{kraemer20_stabl_implem_probab_ode_solver}.
Reference solutions are computed with SciPy \citep{2020SciPy}.
All experiments were conducted on standard consumer-grade CPU.
Code for the implementation and experiments is publicly available on GitHub\footnote{\url{https://github.com/DDCoan/pn-ode-up}}.

\subsection{Propagating uncertainty through nonlinear systems}
\label{sec:experiments:examples}

\begin{figure*}[t]
  \begin{center}
    \includegraphics[width=\linewidth]{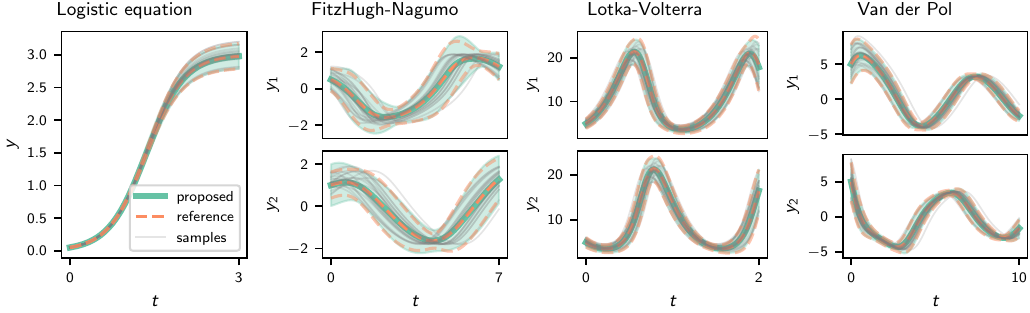}
  \end{center}
  \caption{
    \emph{Propagated uncertainties for the Logistic equation, FitzHugh--Nagumo system, Lotka--Volterra system, and Van der Pol oscillator.}
    The figure shows both the mean (colored lines) and the 95\% credible interval (shaded area)
    computed by our proposed method (``proposed'') and by the reference solution (``reference'') for each system.
    We observe that our method returns mean and variance estimates that largely align with the reference solution.
  }
  \label{fig:performance}
\end{figure*}

\begin{table*}
  \centering
  \caption{
    Uncertain initial value problems used in the experiments.
  }
  \label{tab:ode}
  \begin{tabular}{llll}
    \toprule
    \textbf{Name} & \textbf{Vector field} & \textbf{Initial value} & \textbf{Vector field parameters} \\
    \midrule
    Linear & $f(y, t) = ay + b$ &
    \(\val(0) \sim \N(1, 0.01)\) &
    \(a\!=\!1\), \(b\!=\!0\)
    \\[1em]
    Logistic &  $f(y, t) = ay\!\left( 1 - \frac{y}{b} \right)$ &
    \(\val(0) = 0.05\) & \(a\!=\!3\), \(b\!\sim\!\N(3, 0.01)\)
    \\[1em]
    FitzHugh--Nagumo & $f(t, y) = \begin{bmatrix}y_1 - \frac{1}{3}y_1^3 - y_2 + a \\ \frac{1}{d}(y_1 + b - cy_2)\end{bmatrix}$ &
    \(\val(0) \sim \N(\mqty[0.5\\1], 0.1 \cdot \eye{2})\) &
    \makecell[l]{\(a\!=\!0\), \(b\!=\!0.08\), \(c\!=\!0.07\), \\ \(d\!=\!1.25\)}
    \\[1em]
    Lotka--Volterra & $f(y, t) = \begin{bmatrix} ay_1 - by_1y_2\\ -cy_2 + dy_1y_2\end{bmatrix}$ &
    \(\val(0) \sim \N(\mqty[5\\5], 0.3 \cdot \eye{2})\) &
    \makecell[l]{\(a\!=\!5\), \(b\!=\!0.5\), \(c\!=\!5\), \(d\!=\!0.5\)}
    \\[1em]
    Van der Pol & $ f(y, t) = \begin{bmatrix} y_2 \\ a \cdot (1 - y_1^2)y_2 - y_1 \end{bmatrix}$ &
    \(\val(0) \sim \N(\mqty[5\\5], 2 \cdot \eye{2})\) &
    \(a\!=\!0.05\)
    \\
    \bottomrule
  \end{tabular}
\end{table*}

We apply the proposed uncertainty propagation method to a variety of nonlinear ODE systems, including the Logistic system, the FitzHugh--Nagumo equation, Lotka--Volterra system and the Van der Pol oscillator.
\Cref{tab:ode} provides more information for each system.

To apply our proposed method,
we use an extended Kalman filtering-based ODE filter (also known as EK1) as the underlying ODE solver,
with a once-integrated Wiener process prior
and with automatic calibration of the time-constant diffusion parameter;
all provided by the ProbNum python package \citep{wenger2021probnum}.
We use step sizes of \(10^{-2}\) for the logistic ODE and for Van der Pol, and \(5 \cdot 10^{-3}\) for FitzHugh--Nagumo and Lotka--Volterra.
For numerical quadrature, we use the spherical cubature scheme introduced in
\cref{sec:integration},
which requires \(2d\) applications of the ODE solver for each system, where \(d\) is the dimension of the uncertain parameter space.
The reference solution for each system is computed with Monte Carlo integration over
a classic ODE solver (LSODA as provided by scipy; \cite{2020SciPy}),
using \(100,\!000\) samples and small time steps of \(0.001\).

\Cref{fig:performance} shows the results.
The mean and variance estimates by our probabilistic uncertainty propagation method largely align with the reference solution.
This demonstrates that our proposed method provides reliable uncertainty propagation results across multiple nonlinear systems.
In \cref{sec:appendix}, we show that our proposed method can also deliver reliable uncertainty propagation results for non-Gaussian distributed parameters.

\subsection{Investigating the utility of numerical uncertainty for uncertainty propagation}
\label{sec:experiments:why_pn}

\begin{figure*}[t]
  \begin{center}
    \includegraphics[width=\linewidth]{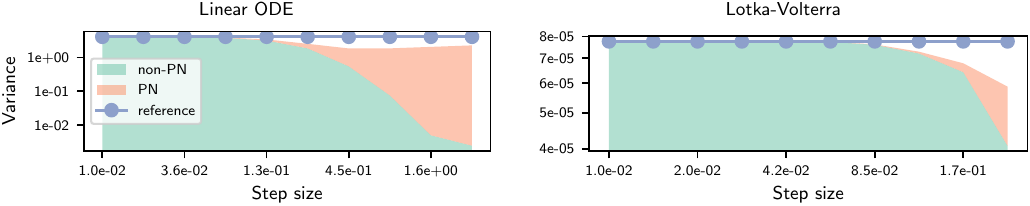}
  \end{center}
  \caption{
    \emph{Decomposition of the propagated variance into its non-PN and PN parts.}
    For both the linear system and the Lotka--Volterra ODE, the non-PN variance decreases for larger step sizes and thereby diverges from the accurate reference variance.
    The PN variance on the other hand, which depends on the numerical uncertainty of the ODE filter, grows for increasing step sizes and helps fill the gap between the non-PN variance and the reference variance to reduce the resulting overconfidence.
  }
  \label{fig:variance_decomposition}
\end{figure*}

In this section, we investigate the utility of the numerical uncertainty estimates provided by the underlying ODE solver.
To this end, we revisit the expression for the computed variance from \cref{eq:mixture_variance}, and decompose it into two parts:
\begin{align}
  \begin{split}
    \V[ \val(t) ]_{\p(\val(t))}
    =&
    \\
    \underbrace{\sum_{i=1}^N w_i \Sigma_i(i)}_\text{PN}
    &+
      \underbrace{
      \sum_{i=1}^N w_i \left( \mu_i(t) - \bar{\mu}(t) \right) \left( \mu_i(t) - \bar{\mu}(t) \right)^\T
      }_\text{non-PN}.
  \end{split}
\end{align}
The non-PN part corresponds exactly to the variance that one would compute without any numerical ODE solver uncertainty (see also \cref{remark:pn-non-pn}).
The PN part on the other hand depends only on the ODE solver uncertainties, and uses these to inflate the resulting total covariance.
In the following, we visualize this variance decomposition to show the utility of the numerical uncertainty estimates provided by the ODE solver.

We consider a linear ODE and a Lotka--Volterra system as described in \cref{tab:ode}, with time spans of \([0, 3]\) and \([0, 0.5]\) respectively,
and with a reduced initial value variance of \(10^{-5} \cdot \eye{2}\) for the Lotka--Volterra problem.
We consider the same ODE filter as in \cref{sec:experiments:examples} for the linear problem and use an ODE filter with twice-integrated Wiener process for the Lotka--Volterra ODE.
The reference variance for Lotka--Volterra is again computed by Monte Carlo integration over
a classic ODE solver (LSODA as provided by scipy; \cite{2020SciPy}),
with \(100,\!000\) samples and small time steps of \(0.001\).

For the linear ODE, we can compute the ground-truth variance analytically.
We then compute the propagated uncertainties for the final time step with our proposed method for 10 different ODE solver step sizes, equally-spaced in log-space in
\([0.01, 3]\) for the linear ODE and \([0.01, 0.25]\) for the Lotka--Volterra ODE, respectively,
and decompose the total variance into the PN and non-PN parts.

\Cref{fig:variance_decomposition} shows the resulting variance decompositions.
On the considered problems, the non-PN part decreases with increasing step sizes and diverges from the accurate reference variance.
But as the numerical error of the ODE solver grows with increasing step sizes, the PN part grows for increasing step sizes and helps fill the gap between the non-PN variance and the reference variance.
This shows that the numerical uncertainty estimates provided by the ODE solver can help to prevent overconfidence in the propagated uncertainty estimates.

\section{Limitations}
The proposed method and the evaluation shown in this paper have several limitations.

\paragraph{Quantification of the quadrature error.}
While we build on probabilistic numerical ODE solvers to quantify errors due to discretization in the time-domain, we do not account for the numerical error due to numerical quadrature.
Bayesian quadrature methods could be used instead to approximate both the integral and the integration error \citep{pnbook2022},
but their application to our problem setting is more involved than simply replacing the numerical quadrature scheme and requires a fully probabilistic treatment of the problem, with joint priors and inference over both time and the ODE parameter.
We leave this for future work.

\paragraph{Higher-order moments and generic expectations.}
Our proposed method only computes the mean and variance of the ODE solution with respect to the parameter distribution, but expectations \(\E[\g(\val(t))]_{\p(\theta)}\) for a general function \(\g\) are also often of interest.
These can in general not be computed analytically for a Gaussian mixture distribution.
One possible approach could be to use a second numerical quadrature scheme to approximate this expectation, but this is left for future work.

\section{Conclusion}
This paper presents a novel approach for propagating uncertainty through filtering-based probabilistic numerical ODE solvers.
While existing probabilistic ODE solvers provide uncertainty estimates for numerical errors, we show that they do not automatically solve the uncertainty propagation problem.
Our method addresses this limitation by combining filtering-based ODE solvers with numerical quadrature to properly marginalize over uncertain system parameters.

The experimental results demonstrate that our approach successfully propagates uncertainty through various nonlinear dynamical systems, with the resulting uncertainty estimates closely matching reference solutions.
Furthermore, we show that the numerical uncertainty estimates provided by probabilistic ODE solvers serve an important role in preventing overconfidence, particularly when using larger step sizes.
This work demonstrates that probabilistic numerical methods can provide meaningful uncertainty quantification not just for numerical errors, but also for uncertainty propagation in dynamical systems.

\section*{Acknowledgements}
NB gratefully acknowledge co-funding by the European Union (ERC, ANUBIS, 101123955. Views and opinions expressed are however those of the author(s) only and do not necessarily reflect those of the European Union or the European Research Council. Neither the European Union nor the granting authority can be held responsible for them). NB thanks the International Max Planck Research School for Intelligent Systems (IMPRS-IS) for their support.
The authors also thank Jonathan Schmidt for helpful feedback.

\bibliography{references}
\clearpage
\appendix
\onecolumn

\section{Additional experiment: Propagating uniform uncertainty through non-linear systems}
\label{sec:appendix}
\begin{figure*}[h!]
  \begin{center}
    \includegraphics[width=\linewidth]{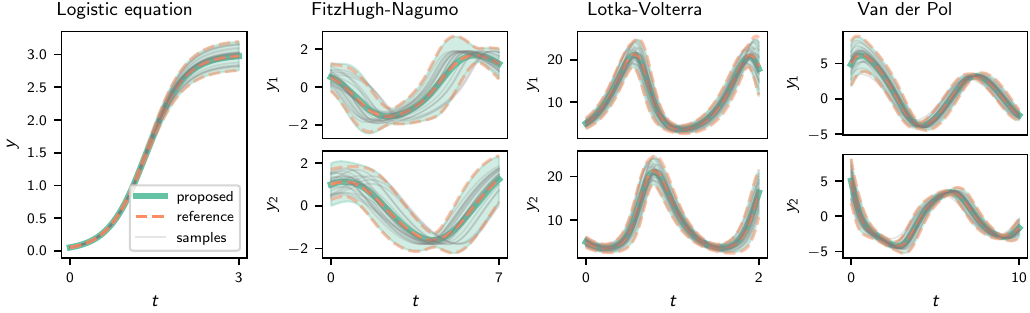}
  \end{center}
  \caption{
    \emph{Propagated uncertainties for the Logistic equation, FitzHugh--Nagumo system, Lotka--Volterra system, and Van der Pol oscillator for non-Gaussian parameters.}
    The figure shows both the mean (colored lines) and the 95\% credible interval (shaded area)
    computed by our proposed method (``proposed'') and by the reference solution (``reference'') for each system.
    We observe that our method returns mean and variance estimates that largely align with the reference solution.
  }
  \label{fig:performance_uniform}
\end{figure*}

\Cref{fig:performance_uniform} provides additional experimental results for uncertainty propagation for uniform parameters.
The parameters are sampled from a uniform distribution $\text{Unif}\,[\mu - 1.96\sigma, \mu+1.96\sigma]$, where $\mu$ and $\sigma$ are chosen as the means and standard deviations of the Gaussian distributions described in \cref{tab:ode}.
The uncertainty is propagated using 1000 Monte Carlo samples.
Other implementational settings follow the details provided in~\cref{sec:experiments}.\\

\end{document}